\documentclass[journal,twocolumn]{IEEEtran}

\usepackage{xcolor,mathtools,subcaption}
\usepackage{algorithm}
\usepackage{algpseudocode}
\usepackage{epsfig,makeidx,color}
\usepackage{amsmath,amssymb,bbm}
\usepackage{cite,graphicx,lipsum}
\usepackage{enumerate}
\usepackage[switch,pagewise]{lineno}
\usepackage{hyperref}
\hypersetup{
        colorlinks = true,
        citecolor=blue,
}
\def\be{ \begin{equation} }
\def\ee{ \end{equation} }
\def\bea{ \begin{eqnarray} }
\def\eea{ \end{eqnarray} }

\def\b0{{\bf 0}}

\begin{document}

\title{Toward Scalable SDN for LEO Mega-Constellations: A Graph Learning Approach} 
\author{Sivaram Krishnan, Bassel Al Homssi, Zhouyou Gu, Jihong Park, Sung-Min Oh, and Jinho Choi

\thanks{
S. Krishnan and J. Choi are with the School of Electrical and Mechanical Engineering, University of Adelaide, Adelaide 5000, Australia (email: \{sivaram.krishnan, jinho.choi\}@adelaide.edu.au)

B. Homssi is with the College of Computing and Informatics, University of Sharjah, Sharjah, United Arab Emirates (email: bhomssi@ieee.org)

Z. Gu and J. Park are with the Singapore University of Technology and Design, Singapore (email: \{zhouyou\_gu, jihong\_park\}@sutd.edu.sg)

S. Oh is with the Electronics and Telecommunications Research Institute, Daejeon, Korea (email: smoh@etri.re.kr)

This study was supported by Institute of Information \& Communications Technology Planning \& evaluation (IITP) grant funded by the Korean government (MSIT) (No. RS-2024-00359235, Development of Ground Station Core Technology for Low Earth Orbit Cluster Satellite Communications).}
}

\maketitle
\begin{abstract}

Terrestrial network limitations drive the integration of non-terrestrial networks (NTNs), notably mega-constellations comprising thousands of low Earth orbit (LEO) satellites. While these satellites act as interconnected network switches via inter-satellite links (ISLs), their massive scale creates severe bottlenecks for network management. To address this, we propose a scalable, hierarchical software-defined networking (SDN) framework. Our architecture leverages graph neural networks (GNNs) to compactly represent the constellation topology, and Koopman theory to linearize nonlinear dynamics. Specifically, a Graph Koopman Autoencoder (GKAE) forecasts spatio-temporal behavior within a linear subspace for each orbital shell. A central SDN controller then aggregates these shell-level predictions for globally coordinated control. Simulations on the Starlink constellation demonstrate that our approach achieves at least a 42.8\% improvement in spatial compression and a 10.81\% improvement in temporal forecasting compared to established baselines, all while utilizing a significantly smaller model footprint.

\end{abstract}

\begin{IEEEkeywords}
mega-constellations; 
Graph Neural Networks (GNN);
Low Earth Orbit (LEO) Satellites; Software Defined Networking (SDN), Koopman Autoencoder
\end{IEEEkeywords}

\section{Introduction}

Thanks to inter-satellite links (ISLs), low Earth orbit (LEO) satellites can form a network that can scale into mega-constellations \cite{Homssi22, Wu25}, enabling seamless interconnection among satellites as they orbit the Earth. Such mega-constellations have the potential to support truly global connectivity with high data rates and low latency, offering significantly improved coverage and service quality even in remote or underserved regions. By leveraging dense satellite deployment and flexible routing through ISLs, these networks can provide resilient and scalable communication services that integrate seamlessly with terrestrial systems as part of non-terrestrial networks (NTNs) \cite{Lin21}, thereby complementing and enhancing existing terrestrial infrastructure.

Software defined networking (SDN) offers centralized, software-based control for networks, providing greater flexibility and scalability than traditional distributed architectures \cite{Kreutz15}. It can align well with the \emph{structured nature} of LEO constellations, which are organized into shells composed of distinct orbital planes whose predictable motion enables accurate modeling of network dynamics and supports the global, time-evolving view required by SDN.


Prior studies show that SDN can improve handover performance \cite{Yang16} and enable more efficient routing through global network awareness \cite{Kumar22}. SDN-based control becomes particularly advantageous in complex scenarios where multiple LEO constellations span different shells or operate across many orbital planes, requiring coordinated management of rapidly changing topology.
In parallel, there has been a growing rise of artificial intelligence (AI) and machine learning (ML) techniques applied to SDN \cite{Xie19} and satellite networking \cite{Krishnan2025SDLEO}. These approaches can enable data-driven optimization, adaptive decision-making, and intelligent control mechanisms that more effectively manage the complexity and dynamics of mega-constellations.

However, managing a large number of LEO satellites presents significant challenges for fully centralized SDN control. While distributed mechanisms, such as end-to-end flow control used in the Internet, can be applied to mega-constellations, they fail to exploit the networks’ highly regular structure and predictable orbital dynamics, which could enable more efficient centralized management. Unlike the Internet, where heterogeneity necessitated distributed control at the cost of efficiency, mega-constellations offer the opportunity to achieve both scalability and efficiency. Addressing these challenges requires efficient representations that leverage the regular structure and dynamics of the constellation, along with learning in a low-dimensional latent space to support effective centralized control and resource management.

In this article, we address the challenge of effectively representing and managing satellite mega-constellations. By treating a mega-constellation as a dynamic graph—where nodes represent LEO satellites and edges represent ISLs—we employ graph neural networks (GNNs) \cite{Scarselli09, Huang25} as a foundational tool to learn topological structures. However, relying solely on standard GNNs is insufficient to manage the massive scale and complex, time-varying dynamics of multi-orbit networks. To overcome this, we propose a hierarchical learning framework designed to equally address both spatial and temporal scalability.

First, to achieve \emph{spatial scalability}, we decompose the global constellation into distinct orbital shells. This shell-specific representation drastically reduces the dimensionality of the state space, allowing localized topologies to be processed independently.

Second, to achieve \emph{temporal scalability}, we model the evolution of each individual shell using a dedicated Graph Koopman Autoencoder (GKAE) \cite{Krishnan2025Koopman}. At its core, Koopman operator theory offers a powerful mathematical paradigm: it posits that any complex, non-linear dynamical system can be equivalently represented by a linear operator acting on a higher-dimensional space of observable functions.  Building on this, the GKAE merges the spatial compression of GNNs with Koopman theory to dynamically 'lift' (to a higher-dimension) the highly non-linear, discrete network states into a continuous, linear latent space. This linearization is the key to temporal scalability. By transforming a non-linear control problem into a linear one, the framework bypasses the compounding errors of traditional recurrent models, enabling long-term predictions that are mathematically stable, highly interpretable, and computationally efficient.

Within this dual-focused framework, we envision a logically centralized controller that aggregates the compact, linearized embeddings from each shell's GKAE, including their forecasted state trajectories.  Using these predictive representations, the controller coordinates inter-shell communication and establishes a pathway to optimize control sequences directly within the latent space, ensuring global coherence.

\section{Software Defined Satellite Networking: Opportunities and Challenges}
\begin{figure}[thb]
\centering
\includegraphics[width=0.9\columnwidth]{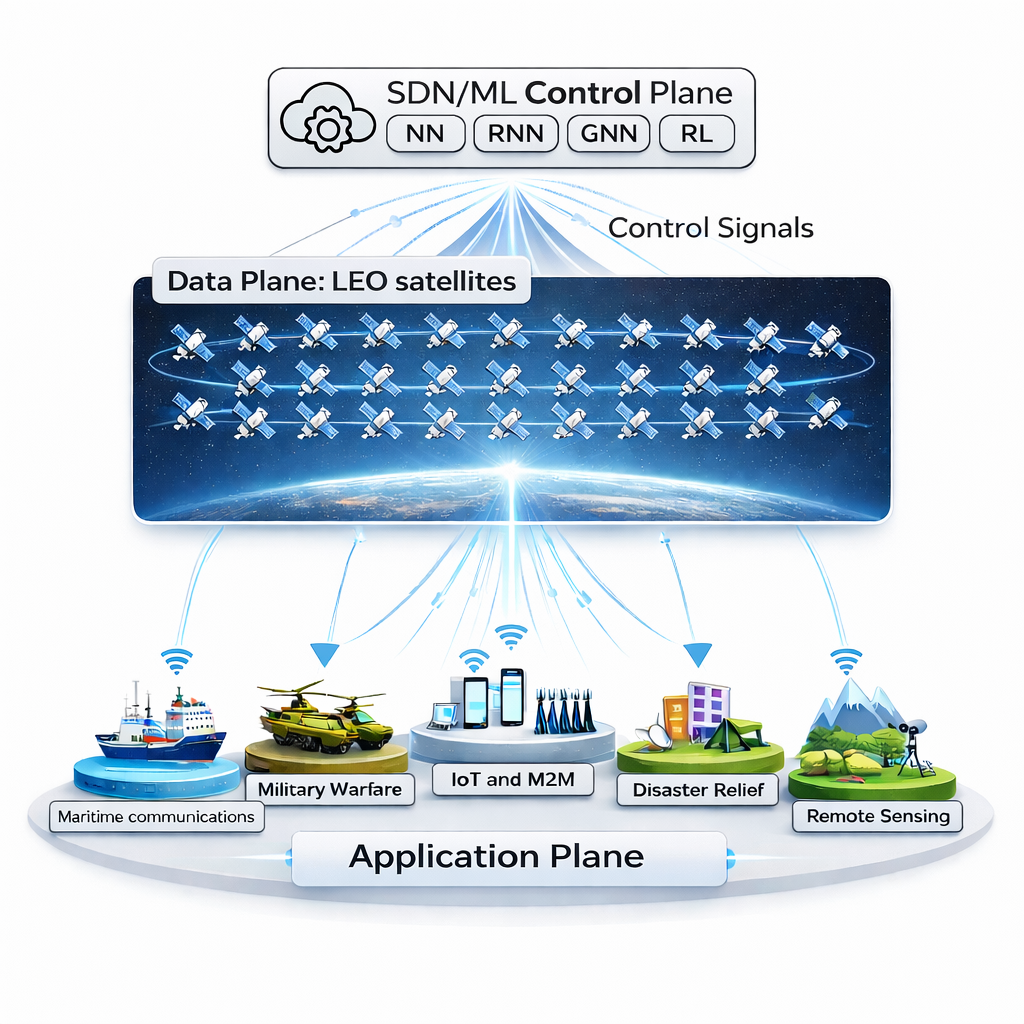}
\caption{An illustration of the SDSN framework. The \textbf{Data Plane} consists of LEO satellites acting as forwarding switches. The \textbf{Control Plane} centralizes network intelligence for effective control and management. The \textbf{Application Plane} defines high-level network policies. (Generative AI was used for editing this schematic)}
    \label{Fig:sdsn}
\end{figure}

\textbf{Software Defined Satellite Networking}: As illustrated in Fig.~\ref{Fig:sdsn}, the software defined satellite networking (SDSN) framework adapts the traditional terrestrial SDN paradigm—decoupling the control plane (decision logic) from the data plane (packet forwarding)—to the dynamic environment of space. In this architecture, LEO satellites function as programmable switches in the data plane, forwarding traffic based on flow tables populated by a logically centralized controller via a Southbound Interface (e.g., space-adapted OpenFlow \cite{mckeown2008openflow}).  To ensure resilience, these control channels can be mapped out-of-band using dedicated telemetry frequencies, or multiplexed in-band over ISLs.

To circumvent the strict size, weight, and power (SWaP) constraints of space hardware, foundational architectures like OpenSAN \cite{bao2014opensan} traditionally host the management plane on the ground. However, relying on ground stations for continuous global orchestration introduces debilitating ground-to-space control loop latency, fundamentally limiting the network's responsiveness.

\textbf{Driving Applications}: Operating through a Northbound Interface, the application plane translates high-level business logic into executable network policies. The core advantage of SDSN is its ability to multiplex diverse, mission-critical services over a single physical infrastructure. Through a central controller, a mega-constellation can dynamically provision secure, low-latency links for defense operations (e.g., Starshield), deliver fiber-like maritime broadband (e.g., Starlink Maritime), and prioritize the rapid downlink of time-critical remote sensing data.

\textbf{Challenges in SDSN}: These diverse applications—from broadband connectivity to real-time sensing—underscore the need for \emph{scalable, adaptive, and high-fidelity} network control:
\begin{itemize}
    \item \emph{The Curse of Dimensionality}: As mega-constellations scale to tens of thousands of satellites, the global state space explodes. This renders traditional, centralized routing optimizations NP-hard and computationally infeasible. Managing this requires a hierarchical, learnable framework capable of compressing immense topological data into a tractable, lower-dimensional space, while accounting for the distinct orbital mechanics and traffic loads of different shells.

    \item \emph{Reactive vs. Predictive Control}: Legacy SDN protocols are inherently reactive, triggering route updates only after congestion or link failures occur. In space, this introduces unacceptable control loop latency. Fortunately, satellite mobility is governed by strict, deterministic orbital mechanics. This inherent predictability demands a shift toward proactive control frameworks—models that leverage learned dynamics to pre-compute routing and resource allocation long before topological shifts degrade performance.
\end{itemize}

\section{A GNN-based Framework for Mega-Constellations}

Wireless networks, including LEO mega-constellations, naturally form a topological structure which consists of various nodes and edges, typically represented as a graph. Unlike terrestrial networks with a fixed infrastructure, each LEO constellation is organized into several \emph{orbital shells}, and each shell consists of multiple \emph{orbital planes}. Each orbital plane contains a number of satellites following the same orbital path, arranged according to a Walker Delta or Star constellation. This arrangement is symmetric: every satellite within a shell experiences identical local connectivity and physical relationships as they operate in the same altitude and inclination. Edges in the graph correspond to ISLs, which typically connect a satellite to its nearest neighbors within the same orbital plane and, in some cases, to satellites in adjacent planes.
\subsection{Graph Construction for an Orbital Shell}
\begin{figure}[thb]
\centering
\includegraphics[width=1\columnwidth]{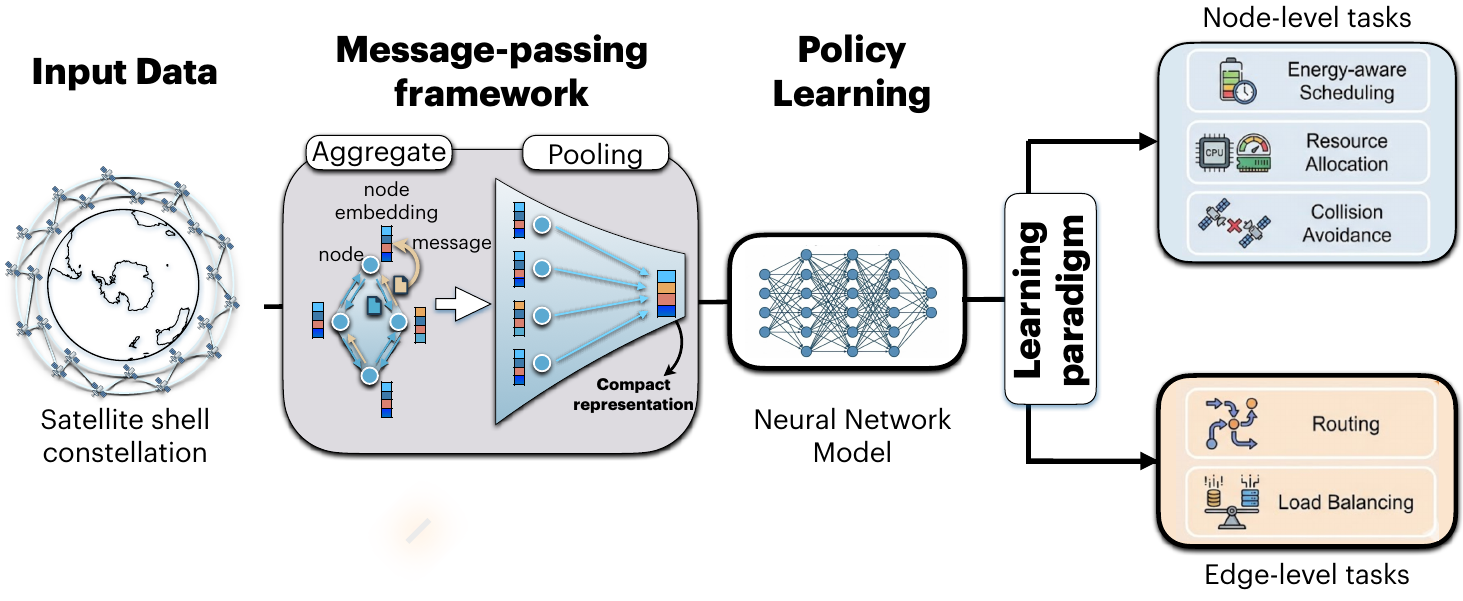}
\caption{An abstraction of satellite networks represented as a compact graph embedding vector.}
    \label{Fig:com}
\end{figure}

We define the graph components to align with the physical satellite system within the same shell, as depicted in Fig.~\ref{Fig:com}. Graphs can be represented using $G = (V, E, X, W)$, which represent the set of nodes, edges, node features and edge weights, respectively. 
\begin{itemize}
    \item \emph{Nodes} ($V$): Each node represents the satellite itself. In an orbital-shell, the set of nodes are all of the satellites within the shell. 
    \item \emph{Edges} ($E$): Edges represent the ISLs, which pertain to the communication pathways between satellites (laser or RF), allowing transmission of packets between satellites. 
    \item \emph{Node Features} ($X$): Node features pertain to the state of a satellite, which could include the spatial information (such as the latitude, longitude and altitude), orbital information (inclination, plane ID) and network state (buffer size, battery level, processing load). 
    \item \emph{Edge Features} ($W$): This attributes to the ISL between two satellites. Key features include the link distance, capacity and the link state (such as active/inactive, signal-to-noise ratio (SNR)). In a shell, there are different types of ISL connections:
        \begin{itemize}
        \item \textit{Intra-plane} edges: These are static links connecting a satellite to its immediate neighbors (successor and predecessor) in the orbital plane. 
        \item \textit{Inter-plane} edges: These are dynamic links connecting a satellite to neighbors in adjacent orbital planes. 
    \end{itemize}
\end{itemize}

\subsection{Graph Neural Network: Learning on Non-Euclidean Data}
Standard convolutional neural networks (CNNs) are designed for structured, grid-like data such as images, which makes them ill-suited for the irregular, non-Euclidean topology of satellite networks. On the other hand, GNNs extend convolution to graphs through message passing, allowing satellites to collaborate rather than operate in isolation. This approach offers several key advantages:
\begin{itemize}
    \item \emph{Permutation Invariance:} The message passing framework is applied simultaneously across all the nodes in the network. Thus, the output node embedding remains the same even if the nodes in the graph are reordered. 
    \item \emph{Inductive Generalization:} GNNs are adaptable for varying network sizes, which is useful for constellations that are deployed in phases. Because the model learns local connectivity patterns rather than a fixed global topology, a GNN trained on a small-scale simulation can be directly applied for inference on a much larger orbital shell without the need for retraining.
    \item \emph{Compression:} GNNs have demonstrated remarkable success in scaling to massive graphs in fields such as social media analysis, wireless networks and computational chemistry. In satellite networks, where the node features are highly correlated among neighbors, GNNs have the potential to exploit this local spatial redundancy to produce significantly more compact representations than traditional methods. 
    This advantage is particularly valuable in SDSN, where a central controller manages the network at a graph level rather than dealing with individual satellites. Instead of tracking each satellite or enumerating all possible links, the controller can operate on the graph representation, reducing computational complexity while preserving the essential structure of the network.    
\end{itemize}

GNNs operate using the message-passing framework, that include three key phases:
\begin{itemize}
    \item \emph{Messages}: Each satellite receives state information ("messages") from its connected neighbors.
    \item \emph{Aggregation}: The satellite aggregates these incoming messages using a permutation-invariant function. 
    \item \emph{Update}: The satellite updates its hidden state based by combining its current features with the aggregated information from its neighbors. Using multiple-layers, the satellite gets to see its $k$-hop neighborhood to understand broader network connection. Additionally, a pooling layer or readout layer can be applied for much more compact representation, providing the global view of the graph. 
\end{itemize}

\subsection{GNN-Driven Scalability for SDSN Control} 
The centralized control paradigm in the SDSN framework, while architecturally useful in providing a global view for optimization of the network, faces a critical implementation bottleneck in the context of mega-constellations due to the curse of dimensionality. 

\begin{figure*}[thb]
\centering
\includegraphics[width=0.8\textwidth]{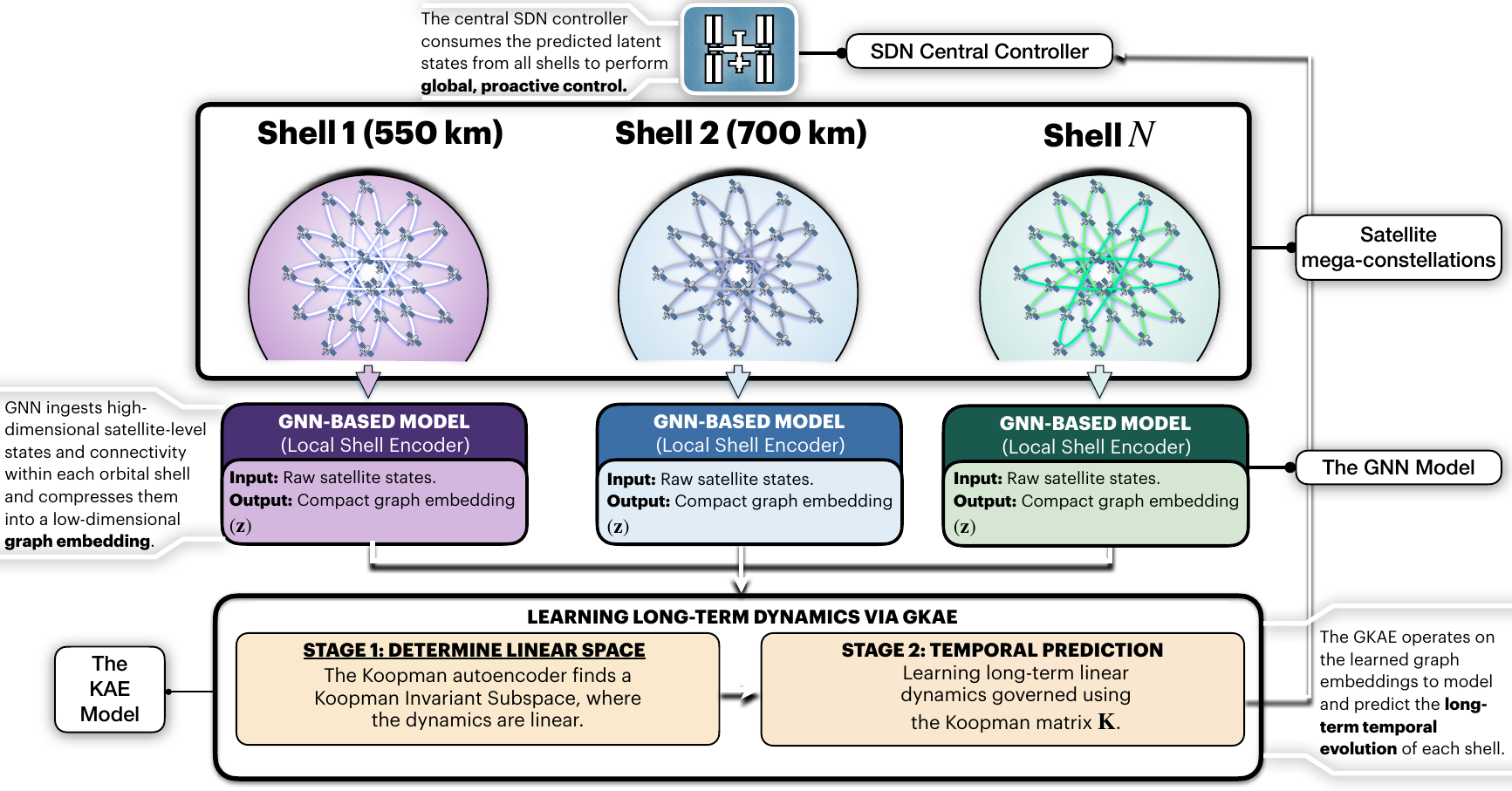}
\caption{Hierarchical Software Defined Satellite Network.}
    \label{Fig:sdsn2}
\end{figure*}

\textbf{Hierarchical SDSN Framework:} Mega-constellations are not monolithic; they are architected as nested series of orbital shells, each defined by a specific altitude and inclination. Since satellites within the same shell exhibit coherent physical dynamics and spatially correlated traffic patterns, the SDSN framework can be designed using a hierarchical structure, as shown in Fig.~\ref{Fig:sdsn2}. The hierarchical framework decomposes the massive global graph into manageable orbital sub-structures. By assigning a GNN-dedicated learner for each orbital shell, we make the problem of learning on mega-constellations computationally feasible and scalable.

Furthermore, this modular approach supports phased deployment of constellations. As new shells are launched to expand the constellation's capacity, corresponding shell-specific models can be instantiated and integrated into the control loop without disrupting the models governing existing active shells. 

\textbf{Functionalities per Shell}: Each orbital shell leverages GNNs to compress the satellite mesh into a compact spatial embedding. However, static compression is insufficient for dynamic, non-linear environments. To forecast the network's evolution, we move beyond heavy, ``black-box" recurrent models (like long short-term memory (LSTM) models) and utilize GKAE \cite{Krishnan2025Koopman}. GKAE posits that non-linear dynamics can be mapped into a linear invariant subspace. Since satellite traffic and trajectories are quasi-periodic, this linearization offers a simpler, interpretable model capable of stable long-term prediction via spectral decomposition. 

This enables a streamlined central controller: instead of processing raw high-dimensional telemetry for several orbits, the central controller aggregates these compressed, linearized predictions from each shell. This architecture yields a ``logically centralized, physically distributed" paradigm, ensuring global optimality with minimal computational overhead.

\textbf{Validation of the Shell-Specific Learner}: The viability of this hierarchical framework rests on a critical premise that a single learning module can effectively compress and predict the dynamics of a massive, dense orbital shell. This will be discussed in Section~\ref{S:Case}.


\section{Case Studies}  \label{S:Case}

For our case study, we consider the largest Starlink shell, consisting of 1442 Starlink satellites at an altitude of 550 km and inclination of 53$^\circ$. For constructing the graph, we use the location coordinates and the queue length of the satellites as the node features for each satellite. For our experiments, we consider case studies as follows.  

\subsection{Case Study 1: Compact Representation and Generalization} The first case study represents how well different ML architectures can effectively represent the data in a lower-dimension. In this case, along with GNN, we also compare the performance of DNN (or MLP) and the CNN architecture in representing the graph as a lower-dimensional variable and reconstructing it. 

\begin{figure}[h]
    \centering
        \centering
        \includegraphics[width=\columnwidth]{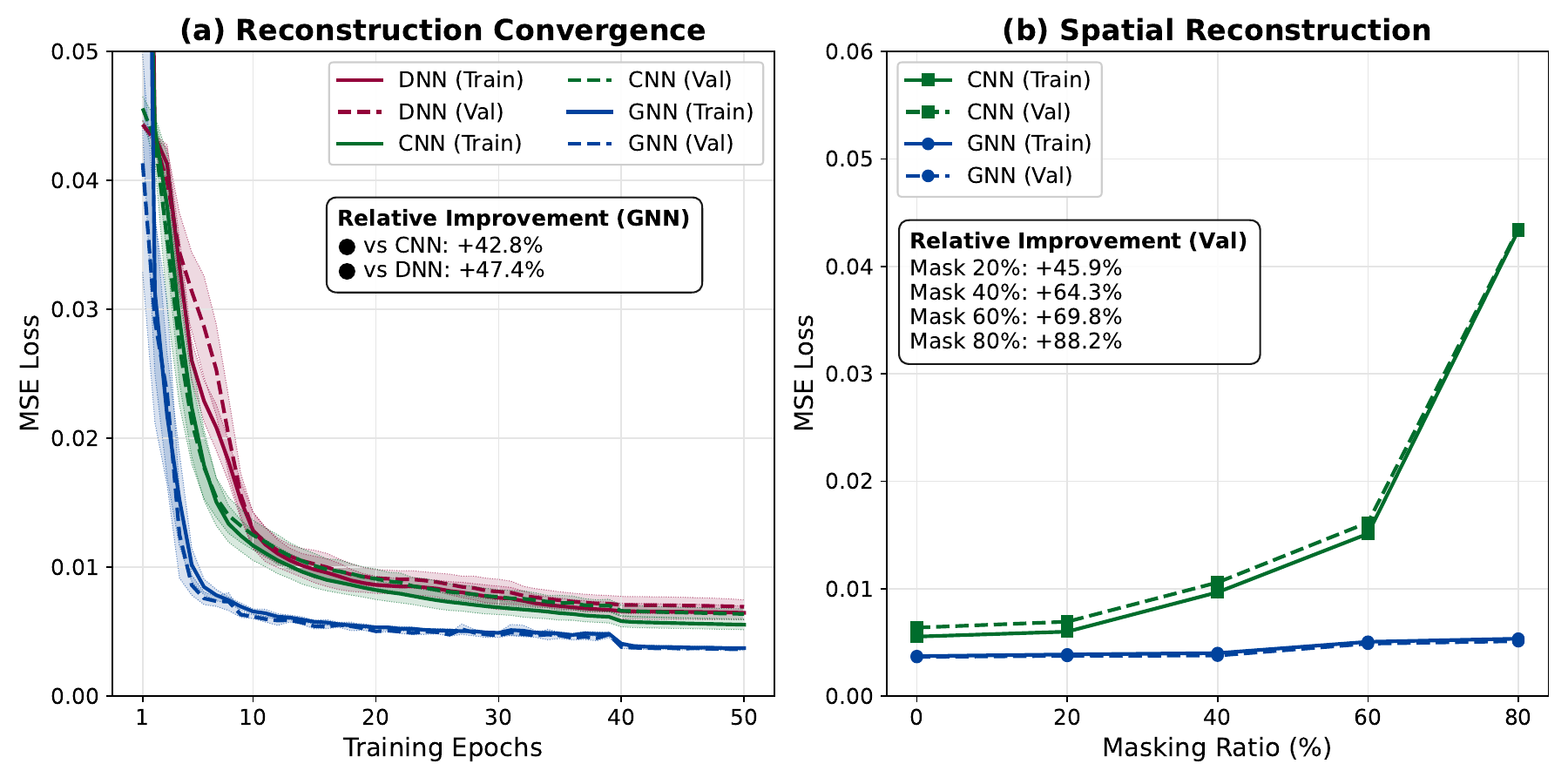}
        \caption{\textbf{Compression and Generalization Analysis}: Compact representation and generalization: (a) Training convergence of varying ML architectures (DNN vs CNN vs GNN) in achieving feature compression and reconstruction, (b) Spatial reconstruction over varying proportion of masking data (CNN vs GNN).}
        \label{fig:t1}
\end{figure}

As seen in the reconstruction task in Fig.~\ref{fig:t1} (a), the GNN achieves a 43\%, 47.4\% improvement against CNN and DNN architectures. It can also be seen that the GNN model generalizes well on the validation set, while being 2.7$\times$ and 3.0$\times$ smaller model than CNN and DNN, respectively. Similarly, in Fig.~\ref{fig:t1} (b), we perform a masked reconstruction, where again, we see that using the message passing framework, the GNN architecture is superior in learning a compact and generalized representation over the satellite networks, achieving $45.9\% - 88.2\%$ lower reconstruction error over the validation set for increasing masking rates.  

\begin{figure}[!h] 
    \centering
    \begin{subfigure}[b]{\columnwidth}
        \centering
        \includegraphics[width=\textwidth]{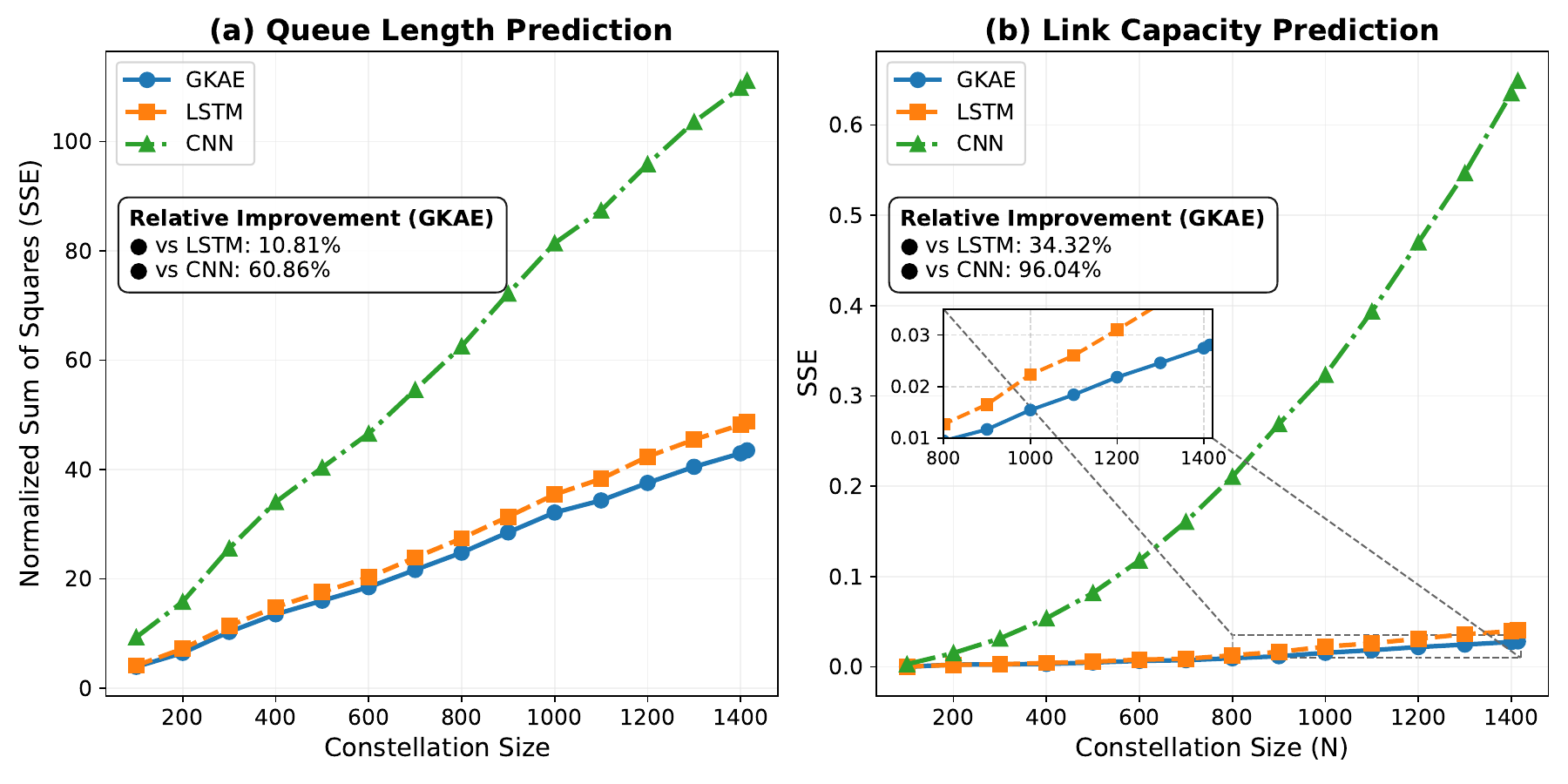}
        \caption{Prediction Accuracy: (left) Queue length prediction and (right) Link Capacity prediction errors. The average improvement in achieving a lower SSE is reported compared to the baseline methods.}
        \label{fig:accuracy}
    \end{subfigure}
    
    \vspace{0.3cm} 
    
    \begin{subfigure}[b]{\columnwidth}
        \centering
        \includegraphics[width=\textwidth]{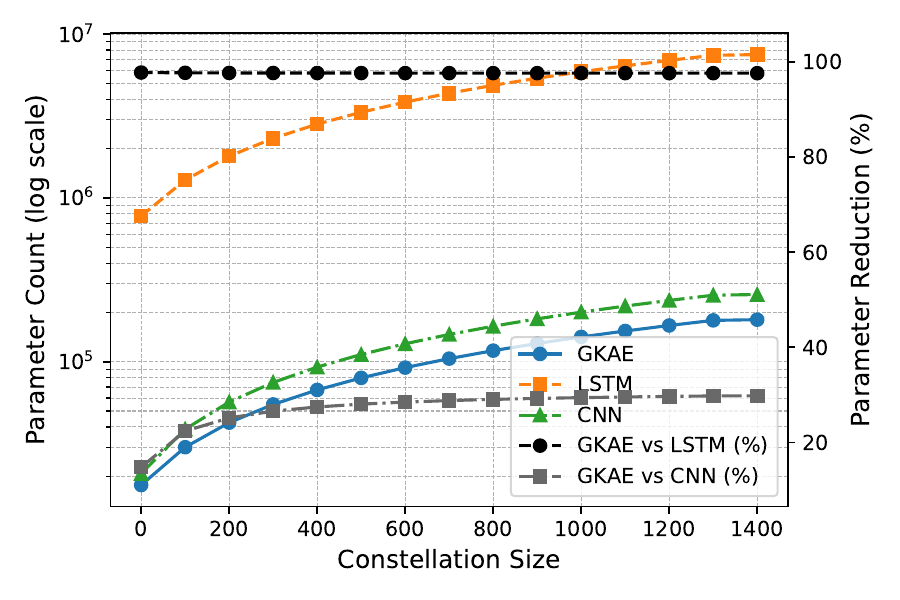}
        \caption{Computational Efficiency: Model Size (Parameters).}
        \label{fig:efficiency}
    \end{subfigure}
    
    \caption{\textbf{Scalability Analysis:} Comparison of prediction accuracy (top row) and computational cost (bottom row) across GKAE, CNN, and LSTM models.}
    \label{fig:full_comparison}
\end{figure}

\subsection{Case Study 2: Time-varying representation and Dynamic Forecasting} For validating the model's ability to capture the non-linear evolution of the network, we evaluate the GKAE on two distinct prediction tasks over a 20-minutes prediction horizon ( sampled at 1-minute intervals): 
\begin{itemize}
    \item Queue length prediction (Traffic Dynamics): This task involves forecasting the buffer occupancy of every satellite in the shell. Accurate multi-step prediction of queue lengths acts as a proxy for anticipating network congestion. In Fig.~\ref{fig:full_comparison}~(left), we compare the predictive performance for the traffic prediction where the GKAE outperforms the LSTM and CNN by 10.81\% and 60.86\%, while being a smaller model. 
    \item Link state prediction: This task focuses on forecasting the future communication availability between satellite (via spectral efficiency forecast), which can help in resource management and routing applications. In Fig.~\ref{fig:full_comparison}~(right), we see a similar trend where the GKAE outperforms these baseline methods over all settings. 
\end{itemize}

In summary, for both case studies, the spatial representation and the temporal forecasting, using a GNN-based framework is suitable over other traditional ML architectures, providing a better reconstruction and prediction accuracy. It is worth noting that GNN-based approach is inherently a smaller model, especially in the second case study, where the model is 2.7x smaller than the CNN and 40x smaller than the LSTM model, respectively.  

Standard queue management and routing protocols, such as Backpressure, are predominantly distributed and reactive. While recent reinforcement learning (RL) approaches can optimize long-term stability for tasks like opportunistic routing \cite{Krishnan2025SDLEO}, most conventional RL strategies are model-free and fail to explicitly capture underlying queue dynamics. By utilizing the proposed framework to proactively learn these queue dynamics, we can significantly enhance existing network protocols. Specifically, predictive awareness allows for the refinement of model-free RL methods and enables Backpressure algorithms to stabilize over a longer horizon, thereby establishing a robust queue management protocol that effectively routes traffic and prevents buffer overflows. Furthermore, successful network operation requires satellites to intelligently offload their buffers. By simultaneously forecasting dynamic link capacities, the controller can perform opportunistic scheduling—transmitting data precisely when inter-satellite channel conditions are optimal.

\section{Open Issues}
In this section, we identify and briefly discuss the open issues apart from the aforementioned challenges related to SDSN. 

\subsection{Learning-Based Network Optimization}
The proposed hierarchical SDSN architecture enables multiple learning paradigms tailored to satellite mega-constellations. In particular, GKAE serves as a foundational representation-learning module that transforms high-dimensional, rapidly evolving network states into compact latent dynamics suitable for downstream optimization and control. At the network level, these latent representations are aggregated to enable global orchestration without requiring full visibility of raw satellite states. This abstraction enables a range of learning-driven control functions, which can be broadly categorized into node-level and edge-level optimization as follows.

\begin{table*}[t]
\centering
\caption{Trade-offs in Central Controller Placement for SDSN Frameworks}
\label{tab:controller_comparison}
\renewcommand{\arraystretch}{1.3} 
\begin{tabular}{p{0.14\linewidth}||p{0.26\linewidth}|p{0.26\linewidth}|p{0.26\linewidth}}
\hline
\textbf{Feature} & \textbf{GEO Satellites (Hierarchical)} & \textbf{Ground Stations (Centralized)} & \textbf{Hybrid (Distributed Space-Ground)} \\ \hline \hline
\textbf{Primary Advantage} & \textbf{Global Visibility:} Fixed relative to Earth, maintaining continuous Line-of-Sight (LoS) with underlying LEO shells for a persistent view of network topology. & \textbf{Computational Supremacy:} Leverages terrestrial hyperscale cloud infrastructure to execute highly complex, non-convex global optimization and traffic engineering. & \textbf{Balanced Trade-off:} Offloads heavy, long-term computations to terrestrial nodes while delegating time-sensitive, localized routing decisions to space-borne nodes. \\ \hline
\textbf{Primary Drawback} & \textbf{Latency \& SWaP Limits:} High propagation delay ($\approx 240$ms RTT) and strict Size, Weight, and Power constraints severely restrict onboard processing capacity. & \textbf{Fragmented Connectivity:} Suffers from intermittent visibility, atmospheric attenuation, and expansive "ocean gaps," creating transient control blind spots. & \textbf{State Synchronization Overhead:} Requires complex, delay-tolerant protocols to maintain state consistency across varying latencies, increasing the risk of race conditions. \\ \hline
\textbf{Architectural Complexity} & \textbf{Moderate:} Demands specialized radiation-hardened hardware and highly efficient, lightweight control algorithms to operate within SWaP limits. & \textbf{Low to Moderate:} Utilizes standard terrestrial SDN protocols but relies heavily on scheduled feeder links and handover management. & \textbf{High:} Necessitates sophisticated distributed control planes, dynamic controller delegation, and robust split-brain resolution mechanisms. \\ \hline
\textbf{Ideal Use Case} & Continuous connectivity outweighs the need for raw processing power. & Deep packet inspection, long-term traffic forecasting, global resource allocation. & Large-scale, dynamic SDSN constellations requiring both rapid local reactions (e.g., link failure recovery) and complex global routing. \\ \hline \hline
\end{tabular}
\end{table*}

\textbf{Node-Level Optimization:} Satellites in mega-constellations act as resource-constrained edge nodes subject to stringent Size, Weight, and Power (SWaP) constraints. Node-level learning focuses on autonomous, predictive management using latent state forecasts, enabling energy-aware scheduling. For instance, GKAE-based trajectory-aware prediction enables forecasting of eclipse cycles and battery dynamics, supporting proactive task scheduling under energy constraints.
Furthermore, onboard resource management is another key application. Latent spatiotemporal traffic representations allow satellites to anticipate congestion and trigger load redistribution or packet rerouting prior to buffer overflow, thus avoiding packet loss. 
Lastly, it enables efficient collision avoidance. Learned dynamical models can support predictive collision risk assessment and fuel-efficient maneuver planning without disrupting active ISLs. 

\textbf{Edge-Level Optimization:} ISLs are the backbone of the constellation but are volatile due to Doppler shifts and Earth blockage. Learning tasks here focus on maximizing link utility. By anticipating future link quality, routing decisions can be proactively made before degradation occurs, thereby reducing the overhead of reactive link-state protocols. In parallel, jointly predicting traffic demand and gateway visibility enables load balancing across orbital shells and planes, allowing traffic to be redistributed preemptively to mitigate congestion and packet loss.

\subsection{Placement of Central Controller}
The physical placement of the central SDSN controller presents a fundamental trade-off between global visibility and computational capability. As summarized in Table~\ref{tab:controller_comparison}, GEO-based controllers offer continuous line-of-sight to LEO shells but suffer from SWaP constraints and inherent propagation latency due to the roughly 30,000 km distance to the LEO orbital plane. Conversely, ground-based controllers provide substantial computational resources but experience intermittent connectivity.  To resolve this, we envision a split-control architecture: computationally intensive model training and global optimization are performed on the ground, while GEO-based controllers leverage the trained, lightweight GKAE models to execute proactive inference and predictive inter-shell coordination. By forecasting network states, this architecture effectively masks the propagation delays that render strict real-time control from GEO physically infeasible.

\subsection{Federated Constellation Learning:}
Federated constellation learning (FCL) can facilitate the coordination between the different shells across the constellation by enabling short model update bursts. Hence, limited resources such as power and bandwidth can be optimized by solely relying on the transmission of few bytes that correspond to model weight updates. This in turn avoids the continuous transmission of the data thus preserving the integrity of the network. Moreover, FCL scales well with large number of nodes as model convergence becomes more feasible making it a feasible approach for mega satellite constellations.

\subsection{Connection with Deep Reinforcement Learning (DRL)}
In order to complement GNNs and to attain robust and resilient decision-making, DRL needs to be integrated~\cite{sutton1998reinforcement}. Agents can learn complex representations that are produced via the GNN to predict the network performance evolution over time without the need for high computational resources. Moreover, this enables delegation across the model and reduces any form of hallucination that may occur via the learning process which enables the ability to close the loop. This is essential where the DRL agents can forecast future states and reduce any potential congestion or bottlenecks across the network.

\subsection{Towards Heterogeneous Graphs in 6G NTN:}
LEO satellite networks were never envisioned to operate as stand-alone networks, but rather to complement the currently deployed terrestrial infrastructures as well as other non-terrestrial networks. In order to facilitate this integration, the proposed framework needs to extend to enable \emph{heterogeneous graphs} (HeteroGNNs). HeteroGNNs enable nodes to possess different capabilities through the learning of a unified space via a main node or controller. This ensures seamless integration and operations of the heterogeneous network. This may be realized through offloading the LEO network traffic to ground or GEO relays or the autonomous deployment of UAVs or high altitude platforms (HAPs) to coverage hole geo-locations.

\subsection{Future NTN and Space O-RAN} Finally, the proposed hierarchical framework aligns naturally with the emerging Open radio access network (O-RAN) architecture for NTN \cite{baena2025space}. O-RAN standards split network management into two distinct layers: ``Near-Real-Time" control for fast, local decisions, and ``Non-Real-Time" orchestration for global, slower-paced policy training.

Our GNN-based hierarchy mirrors this standardized split exactly. The shell-level learners act as Near-Real-Time controllers, handling rapid inference for local routing and resource allocation within the shell. Meanwhile, the central SDSN controller functions as the Non-Real-Time orchestrator, aggregating global data to retrain models and update long-term policies. This structural alignment suggests that GNN-based SDSN is not just a theoretical model, but a feasible software module compatible with future standardized 6G architectures.

\section{Conclusion}
In this article, we explore the emerging challenge of operating satellite mega-constellations using SDN principles. Traditional reactive control mechanisms struggle to cope with thousands of fast-moving satellites, volatile ISLs, and intermittent ground connectivity. To address these challenges, we proposed a proactive hierarchical SDSN framework that decomposes the global constellation into orbital shells to enable scalable control and coordination. A central theme of this framework is the use of learning-based predictive abstractions to compress high-dimensional constellation dynamics into compact representations that are suitable for real-time orchestration. The GKAE is used as a model for learning the dynamics for each shell constellation. In our case studies, we illustrate how shell-level dynamics can be learned, forecasted, and aggregated at the central controller to support proactive decision-making with reduced signaling and latency overhead.

\bibliographystyle{ieeetr}
\bibliography{LEO}

\end{document}